# Fast Exact Inference for Recursive Cardinality Models

**Daniel Tarlow, Kevin Swersky, Richard S. Zemel,**
Dept. of Computer Science
University of Toronto
{dtarlow,kswersky,zemel}@cs.toronto.edu

**Ryan P. Adams,**
Sch. of Eng. & Appl. Sci.
Harvard University
rpa@seas.harvard.edu

**Brendan J. Frey**
Prob. & Stat. Inf. Group
University of Toronto
frey@psi.toronto.edu

## Abstract

Cardinality potentials are a generally useful class of high order potential that affect probabilities based on how many of $D$ binary variables are active. Maximum a posteriori (MAP) inference for cardinality potential models is well-understood, with efficient computations taking $\mathcal{O}(D \log D)$ time. Yet efficient marginalization and sampling have not been addressed as thoroughly in the machine learning community. We show that there exists a simple algorithm for computing marginal probabilities and drawing exact joint samples that runs in $\mathcal{O}(D \log^2 D)$ time, and we show how to frame the algorithm as efficient belief propagation in a low order tree-structured model that includes additional auxiliary variables. We then develop a new, more general class of models, termed *Recursive Cardinality models*, which take advantage of this efficiency. Finally, we show how to do efficient exact inference in models composed of a tree structure *and* a cardinality potential. We explore the expressive power of Recursive Cardinality models and empirically demonstrate their utility.

## 1 Introduction

Probabilistic graphical models are widely used in machine learning due to their representational power and the existence of efficient algorithms for inference and learning. Typically, however, the model structure must be restricted to ensure tractability. To enable efficient *exact* inference, the most common restriction is that the model have low tree-width.

A natural question to ask is if there are other, different restrictions that we can place on models to ensure tractable exact or approximate inference. Indeed, a celebrated result is the ability of the "graph cuts" algorithm to exactly find the maximum a posteriori (MAP) assignment in any pairwise graphical model with binary variables, where the internal potential structure is restricted to be submodular. Along similar lines, polynomial-time algorithms can exactly compute the partition function in an Ising model if the underlying graph is planar (Fisher, 1961).

Extensions of these results have been a topic of much recent interest, particularly for the case of MAP inference. Gould (2011) shows how to do exact MAP inference in models with certain higher order terms via graph cut-like algorithms, and Ramalingham et al. (2008) give results for multilabel submodular models. Tarlow et al. (2010) provide efficient algorithms for a number of other high-order potentials.

Despite these successes in finding the optimal configuration, there has been relatively less progress in efficient high order marginalization and sampling. This partially stems from the difficulty of some of the computations associated with summation in these models. For example, computing the partition function for binary pairwise submodular models (where graph cuts can find the MAP) is #P-complete, so we do not expect to find an efficient exact algorithm.

One important high-order potential where such hardness results do not exist is the cardinality potential, which expresses constraints over the number of variables that take on a particular value. Such potentials come up in natural language processing, where they may express a constraint on the number of occurrences of a part-of-speech, e.g., that each sentence contains at least one verb. In computer vision, a cardinality potential might encode a prior distribution over the relationships between size of an object in an image and distance from camera. In a conference paper matching system, cardinality potentials could enforce a requirement that e.g. each paper have 3-4 reviews and each reviewer receive 8-10 papers.

A simple form of model containing a cardinality potential is a model over binary variables, where the model probability is a Gibbs distribution based on an energy function consisting of unary potentials $\theta_d$ and one cardinality potential $f(\cdot)$:

$$-E(\boldsymbol{y}) = \sum_d \theta_d y_d + f(\sum_d y_d) \qquad (1)$$

$$p(\boldsymbol{y}) = \frac{\exp\{-E(\boldsymbol{y})\}}{\sum_{\boldsymbol{y'}} \exp\{-E(\boldsymbol{y'})\}}, \qquad (2)$$

where no restrictions are placed on $f(\cdot)$. We call this the *standard cardinality potential model*. Perhaps the best-known algorithm in machine learning for computing marginal probabilities is due to Potetz and Lee (2008); however, the runtime is $\mathcal{O}(D^3 \log D)$, which is impractical for larger problems.

We observe in this paper that there are lesser-known algorithms from the statistics and reliability engineering literature that are applicable to this task. Though these earlier algorithms were not presented in terms of a graphical modeling framework, we will present them as such, introducing an interpretation as a two step procedure: (i) create auxiliary variables so that the high order cardinality terms can be re-expressed as unary potentials on auxiliary variables, then (ii) pass messages on a tree-structured model that includes original and auxiliary variables, using a known efficient message computation procedure to compute individual messages. The runtime for computing marginal probabilities with this procedure will be $\mathcal{O}(D \log^2 D)$. This significant efficiency improvement over the Potetz and Lee (2008) approach makes the application of cardinality potentials practical in many cases where it otherwise would not be. For example, exact maximum likelihood learning can be done efficiently in the standard cardinality potential model using this formulation.

We then go further and introduce a new high order class of potential that generalizes cardinality potentials, termed Recursive Cardinality (RC) potentials, and show that for balanced RC structures, exact marginal computations can be done in the same $\mathcal{O}(D \log^2 D)$ time. Additionally, we show how the algorithm can be slightly modified to draw an exact sample with the same runtime. We follow this up by developing several new application formulations that use cardinality and RC potentials, and we demonstrate their empirical utility. The algorithms are equally applicable within an approximate inference algorithm, like loopy BP, variational message passing, or tree-based schemes. This also allows fast approximate inference in multi-label models that contain cardinality potentials separately over each label.

Finally, we show that cardinality models can be combined with a tree-structured model, and again assuming a balanced tree, exact inference can be done in the same $\mathcal{O}(D \log^2 D)$ time (for non-balanced trees, the runtime is $\mathcal{O}(D^2)$). This leads to a model class that strictly generalizes standard tree structures, which is also able to model high order cardinality structure.

## 2 Related Work

### 2.1 Applications of Cardinality Potentials

Cardinality potentials have seen many applications, in diverse areas. For example, in worker scheduling programs in the constraint programming literature, they have been used to express regulations such "each sequence of 7 days must contain at least 2 days off" and "a worker cannot work more than 3 night shifts every 8 days" (Régin, 1996). Milch et al. (2008) develop cardinality terms in a relational modeling framework, using a motivating example of modeling how many people will attend a workshop. In error correcting codes, message passing-based decoders often use constraints on a sum of binary variables modulus 2 (Gallager, 1963). Another application is in graph problems, such as finding the maximum-weight $b$-matching, in which the cardinality parameter $b$ constrains the degree of each node in the matching (Huang & Jebara, 2007), or to encode priors over sizes of partitions in graph partitioning problems (Mezuman & Weiss, 2012).

More recently, cardinality potentials have become popular in language and vision applications. In part-of-speech tagging, cardinalities can encode the constraint that each sentence contains at least one verb and noun (Ganchev et al., 2010). In image segmentation problems from computer vision, they have been utilized to encourage smoothness over large blocks of pixels (Kohli et al., 2009), and Vicente et al. (2009) show that optimizing out a histogram-based appearance model leads to an energy function that contains cardinality terms.

### 2.2 Maximization Algorithms

As noted previously, there is substantial work on performing MAP inference in models containing one or more cardinality potentials. In these works, there is a division between methods for restricted classes of cardinality-based potential, and those that work for arbitrary cardinality potentials. When the form of the cardinality potential is restricted, tractable exact maximization can sometimes be performed in models that contain many such potentials, e.g., Kohli et al. (2009); Ramalingham et al. (2008); Stobbe and Krause (2010); Gould (2011). A related case, where maximization can only be done approximately is the "pattern potentials" of Rother et al. (2009). For arbitrary functions of counts, the main approaches are that of Gupta et al. (2007) and Tarlow et al. (2010). The former gives a simple $\mathcal{O}(D \log D)$ algorithm for performing MAP inference, and the latter gives an algorithm with the same complexity for computing messages necessary for max-product belief propagation.

### 2.3 Summation Algorithms

Relatively less work in the machine learning community has examined efficient inference of marginal probabilities in models containing cardinality potentials. The best-known approach is by Potetz and Lee (2008) (here PL). Also related is the line of work including de Salvo Braz et al. (2005) and Milch et al. (2008), but these works assume restrictions on unary potentials, and it is not clear that they are efficient in the case where there are many distinct unary potentials.

PL works with potentials $\theta(\boldsymbol{y}; \boldsymbol{w}) = f(\boldsymbol{y} \cdot \boldsymbol{w})$, where $\boldsymbol{y}$ and $\boldsymbol{w}$ are real-valued vectors. This is a general case that is more involved than cardinality potentials, requiring a clever strategy for representing messages, e.g., with adaptive histograms. However, if $\boldsymbol{y}$ is binary and $\boldsymbol{w}$ is the all-ones vector, then $f(\cdot)$ is a standard cardinality potential.

The starting point for PL is to write down the expression for a sum-product message from $f$ to, say, the last variable $y_D$:

$$m_{fD}(y_D) = \sum_{\boldsymbol{y} \setminus \{y_D\}} \left[ f(\sum_d y_d) \prod_{d' \neq D} m_{d'f}(y_{d'}) \right]. \quad (3)$$

Next, PL defines a change of variables into a new set of integer-valued variables, $\boldsymbol{z}$, as follows:

$$z_1 = y_1 \quad z_2 = z_1 + y_2 \quad \ldots \quad z_{D-1} = z_{D-2} + y_{D-1}.$$

Eq. (3) can then be expressed in terms of $\boldsymbol{z}$ as $\sum_{\boldsymbol{z} \setminus \{z_d\}} \left[ f(y_D + z_{D-1}) \prod_{d' \neq d} m_{d'f}(z_{d'} - z_{d'-1}) \right]$.[1] Finally, the sums can be pushed inwards as in variable elimination, and the internal sums can be computed from inside outwards naively in time $\mathcal{O}(D^2)$. There are $D$ summations to perform, so computing one message takes $\mathcal{O}(D^3)$ time. As observed by Felzenszwalb and Huttenlocher (2004) and noted by PL, the internal sums can be performed via FFTs in $\mathcal{O}(D \log D)$ time. Computing all $D$ messages or marginals, then, requires $\mathcal{O}(D^3 \log D)$ time.

## 3 Other Summation Algorithms

Here, we review lesser-known work from the statistics and reliability engineering literature, where efficient algorithms for very similar tasks have been developed.

The idea of a recursive procedure that sums configurations in a chain-structure i.e. by using a definition of $z$

---
[1] For notational convenience, assume we have $z_0 = 0$.

variables similar to that of PL, dates back at least to Gail et al. (1981). In this work, the algorithmic challenge is to compute the probability that exactly $k$ of $D$ elements are chosen to be on, given that elements turn on independently and with non-uniform probabilities. Naively, this would require summing over $\binom{D}{k}$ configurations, but Gail et al. shows that it can be done in $\mathcal{O}(Dk)$ time using dynamic programming.

A similar task is considered by Barlow and Heidtmann (1984) and Belfore (1995) in the context of reliability engineering. The task considered computing the probability that exactly $k$ elements are chosen to be on, or the probability that between $k$ and $l$ elements are chosen to be on. Belfore gives a divide-and-conquer algorithm that recursively calls an FFT routine, leading to an $O(D \log^2 D)$ algorithm. This algorithm is very similar to the approach we take in this work, and in the case of ordinary cardinality potentials (i.e. not RC potentials), it is equivalent to the upward pass of message passing that we will present in Section 5.

Finally, there is work in statistics on Poisson-Binomial (PB) distributions, which are the marginal distributions over cardinalities that arise if we have a model that contains only unary potentials. These distributions have been used in several applications in the statistics literature. We refer the reader to Chen and Liu (1997) and references therein for more details on applications.

One algorithm for computing the cumulative distribution function (CDF) of the PB distribution that uses an FFT was proposed in Fernandez and Williams (2010) and analyzed by Hong (2011). The idea is to first compute the characteristic function of the distribution, and then directly apply the inverse FFT in order to recover the CDF. The benefit of this approach is that the FFT only needs to be applied once, however computing the characteristic function still takes $\mathcal{O}(D^2)$ time.

For further discussion of other algorithms similar to the ones discussed above, we recommend Hong (2011) and references therein.

## 4 Recursive Cardinality Potentials

### 4.1 Model Structure

The first contribution in this paper is to generalize the structure of the cardinality potential to the Recursive Cardinality potential. Recursive Cardinality potentials are defined in terms of a set of subsets $\mathcal{S}$ of a set of $D$ binary variables. The joint probability over vector $\boldsymbol{y}$ is defined as

$$p(\boldsymbol{y}) \propto \prod_{s_k \in \mathcal{S}} f_k(\sum_{y_d \in s_k} y_d), \quad (4)$$

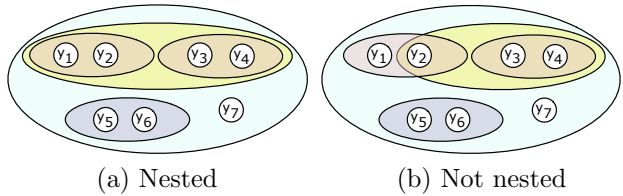

(a) Nested          (b) Not nested

Figure 1: Examples of nested and non-nested subsets.

where the only constraint is that $\mathcal{S}$ is *nested*, as illustrated in Fig. 1. We call a set of subsets $\mathcal{S}$ nested if for every pair of subsets in $\mathcal{S}$, either they are disjoint or one is a subset of the other. Each $f_k$ can be arbitrary, and different per $k$. By defining subsets over single variables, we can represent unary potentials, so we do not explicitly separate them out here.

This new construction extends the standard cardinality potential to handle multiple scales, ranging from purely local (e.g., for a pair of variables) to global, and potentially including all scales in between.

### 4.2 Example Cardinality Potentials

Cardinality potentials can be applied in diverse ways in order to capture a variety of interesting properties. Here we give some examples, and note that this list is far from exhaustive.

#### 4.2.1 Ordinary Cardinality Potentials

**Noisy-OR:** Consider a set of $D$ binary variables $\boldsymbol{y}$ along with a single binary variable $t$ that depends on $\boldsymbol{y}$: $P(t=1|\boldsymbol{y}) = 1-(1-\epsilon)\prod_{d=1}^{D}(1-\lambda_d)^{y_d}$. For the special case where each $\lambda_d$ is equal, we can represent this as a conditional model with a cardinality potential:

$$P(t=1|\boldsymbol{y}) = 1-(1-\epsilon)(1-\lambda)^{\sum_{d=1}^{D} y_d} \quad (5)$$

This is simply a function of $\sum_d y_d$, and can therefore be viewed as a cardinality potential.

**Smoothness and Competition:** Note that even unimodal cardinality potentials have interesting properties. A convex cardinality potential is related to smoothness: it will tend to favor configurations where all of the variables are either on together or off together. In a similar manner, a concave function will cause competition amongst the binary variables.

#### 4.2.2 Recursive Cardinality Potentials

**Group and structured sparsity:** By placing a cardinality potential over subsets of variables consisting of a uniform distribution over counts (or any general distribution) along with a spike at 0, one can represent the preference that variables should tend to turn off together in groups. Indeed, one can represent hierarchical sparsity using a recursive model. This gives an interesting alternative to the traditional approaches of $\ell_0$ and $\ell_1$ priors, as well as their group counterparts that are commonly used in structured sparsity (Zhao et al., 2006).

**Hierarchical CRFs:** A common approach to image segmentation in the computer vision literature is to construct a hierarchy of increasingly coarse segmentations, then to perform the segmentation jointly at the different levels of coarseness. To enforce consistency across levels of coarseness, a widely-used form of potential is the $P^n$ (Kohli et al., 2009) potential, which encourages sets of variables to all take on the same label. If the segmentations at different levels of granularity have a nested structure, then this model can be represented using an RC potential, and thus exact marginal inference (and therefore learning) can be done efficiently.

## 5 Fast Sum-Product Formulation

**Overview.** Here, we present the fast FFT algorithm as an auxiliary variable method, where auxiliary variables are invented in such a way that the distribution over $\boldsymbol{y}$ remains unchanged, but inference in the expanded model can be done very efficiently. In other words, we are defining an augmented model $q(\boldsymbol{y}, \boldsymbol{z})$ that has the property that $\sum_z q(\boldsymbol{y}, \boldsymbol{z}) = p(\boldsymbol{y})$, but where computing all marginals in $q$ (for both $\boldsymbol{y}$ and $\boldsymbol{z}$ variables) can be done more efficiently than directly computing marginals for each $y$ in the original $p(\boldsymbol{y})$ (at least by using any existing method).

More specifically, the algorithm can be described as follows: auxiliary variables $\boldsymbol{z}$ will be integer-valued variables that represent the count over subsets of original variables $\boldsymbol{y}$. The auxiliary variables are structured into a binary tree, as illustrated in Fig. 2, with original $\boldsymbol{y}$ variables at leaves of the tree, so $z$ variables at higher levels of the tree represent sums of increasingly large subsets of $\boldsymbol{y}$. There is one auxiliary variable at every internal binary tree node, so with a balanced tree structure, the joint model over $\boldsymbol{y}$ and $\boldsymbol{z}$ is a tree of depth $\log D$. The utility of this formulation is that we can now represent cardinality potentials over subsets of $\boldsymbol{y}$ variables as unary potentials over $\boldsymbol{z}$ variables.

Having constructed the auxiliary variable model, the algorithm will be simply to run an inwards and outwards pass of sum-product belief propagation. The key computational point is that due to the structure of potentials over auxiliary variables, sum-product messages can be computed efficiently even when $\boldsymbol{z}$ variables have large cardinality, by using FFTs.

### 5.1 Detailed Description

The algorithm takes as input a binary tree $\mathcal{T}$ with $D$ leaf nodes, where each leaf corresponds to one $y_d$

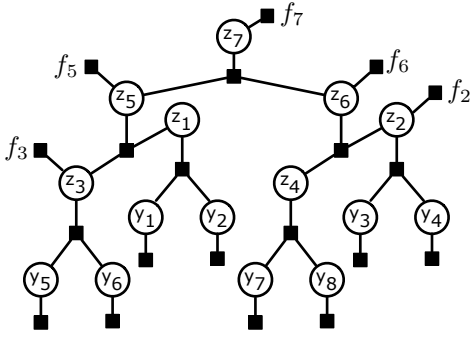

Figure 2: Each internal node $z$ represents the count over the subset of original $y$ variables that are descendants of the internal node. Cardinality potentials on the subset then can trivially be added as unary potentials on internal nodes. Here, in addition to a standard cardinality potential $f_7$, we add subset cardinality potentials $f_2, f_3, f_5,$ and $f_6$.

variable. To instantiate latent variables $\boldsymbol{z}$, traverse up the tree from leaves to root, associating a $z$ variable with each internal node. When instantiating a new variable $z_p$, set a deterministic relationship between it and its two children, $z_l$ and $z_r$, as $z_p = z_l + z_r$. In graphical model terms, for each parent in the tree, add a deterministic potential $g_p(z_p, z_l, z_r) = \mathbf{1}_{\{z_p = z_l + z_r\}}$ to the model. With these definitions, we can define the distribution $q$ as,

$$q(\boldsymbol{y}, \boldsymbol{z}) \propto p(\boldsymbol{y}) \prod_{p \in \mathcal{P}} \mathbf{1}_{\{z_p = z_{l(p)} + z_{r(p)}\}}, \qquad (6)$$

where $\mathcal{P}$ is the set of parent nodes in $\mathcal{T}$, and $l(p)$ and $r(p)$ are indices of the left and right children of $z_p$.

Define the set of leaf node descendants of $z_p$ as $s_p = \{y_d \mid z_p \text{ is an ancestor of } y_d\}$. For any setting of $\boldsymbol{z}$ to have nonzero probability, it clearly must be the case that $z_p = \sum_{y_d \in s_p} y_d$. Thus, a high order potential over subset $s_p$ can equivalently be represented as a unary potential on $z_p$. Expanding the definition of $p$ from within Eq. (6), the computational benefit becomes clear, as we can rewrite all high order potentials as unary potentials:

$$q(\boldsymbol{y}, \boldsymbol{z}) \propto \prod_{k \mid s_k \in \mathcal{S}} f_k(\sum_{y_d \in s_k} y_d) \prod_d \theta_d(y_d) \prod_{p \in \mathcal{P}} \mathbf{1}_{\{z_p = z_{l(p)} + z_{r(p)}\}}$$

$$= \prod_{k \mid s_k \in \mathcal{S}} f_k(z_k) \prod_d \theta_d(y_d) \prod_{p \in \mathcal{P}} \mathbf{1}_{\{z_p = z_{l(p)} + z_{r(p)}\}} \qquad (7)$$

The following proposition justifies correctness and makes the relationship between $p$ and $q$ precise:

**Proposition 1.** *For all $\boldsymbol{y}$, $p(\boldsymbol{y}) = \sum_{\boldsymbol{z}} q(\boldsymbol{y}, \boldsymbol{z})$.*

*Proof.* There is exactly one joint setting of $\boldsymbol{z}$ with nonzero probability for each joint setting of $\boldsymbol{y}$. To see this, observe that given a setting of $\boldsymbol{y}$, the one and only setting of $\boldsymbol{z}$ that satisfies all the deterministic relationships is for each $z_p$ to set $z_p = \sum_{y_d \in s_p} y_d$. The proposition then follows directly. □

### 5.2 Aligning Nested Subsets with $\mathcal{T}$

We have shown how adding auxiliary variables $\boldsymbol{z}$ enables us to convert high order cardinality potentials $f_k(\sum_{y_d \in s_k} y_d)$ into a unary potential on an individual $z_k$ variable. However, this transformation only works if there is an internal $z$ variable for each subset $s_k$ that we wish to put a cardinality potential over. This restriction is what leads to the nested property that we require of sets of subsets $\mathcal{S}$. For any nested set of subsets $\mathcal{S}$, however, it is possible to construct a binary tree $\mathcal{T}$ such that an internal node in the tree is created for every subset $s_k$. A question for future work is whether a weaker condition than nestedness can be enforced while still guaranteeing efficient exact inference.

Finally, all that remains is to show how sum-product messages can be computed efficiently.

#### 5.2.1 Upward Messages

Let $g(z_p, z_l, z_r)$ be a factor that enforces the deterministic relationship $z_p = z_l + z_r$ between parent $z_p$ and children $z_l$ and $z_r$. The upward message vectors that we need to compute take the following form:

$$m_{g,z_p}(z_p) = \sum_{z_l=0}^{c_l} \sum_{z_r=0}^{c_r} g(z_p, z_l, z_r) m_{z_l,g}(z_l) m_{z_r,g}(z_r)$$

$$= \sum_{z_l = z_p - z_r}^{c_l} m_{z_l,g}(z_l) m_{z_r,g}(z_p - z_l).$$

Expressed in this way, it becomes clear that the entire message vector (the above quantity, for all values of $z_p \in \{0, \ldots, c_p\}$) can be computed as a 1D discrete convolution. Since 1D discrete convolutions of vectors of length $N$ can be computed in $\mathcal{O}(N \log N)$ time using FFTs, these message vectors can be computed in $\mathcal{O}(c_p \log c_p)$ time.

#### 5.2.2 Downward Messages

Let $g(z_p, z_l, z_r)$ be a factor that enforces the deterministic relationship $z_p = z_l + z_r$ between parent $z_p$ and children $z_l$ and $z_r$. The downward message vectors that we need to compute take the following form (assume w.l.o.g. that the message is to $z_l$):

$$m_{g,z_l}(z_l) = \sum_{z_r=0}^{c_r} m_{z_p,g}(z_l + z_r) m_{z_r,g}(z_r).$$

This also can be computed as a 1D discrete convolution, after reversing the message vector $m_{z_r,g}$.

### 5.2.3 Asymptotics

Assuming balanced binary trees, the algorithm runs in $\mathcal{O}(D \log^2 D)$ time. This can be seen by using the Master theorem to solve the recurrence $T(n) = 2T(n/2) + n \log n$. If binary trees are not balanced, the worst case runtime can be $O(D^2)$. The algorithm uses $\mathcal{O}(D \log D)$ space.

### 5.3 Minor Extensions

We argue that a main benefit of the auxiliary variable formulation is that it allows for several variants and elaborations. We discuss some of them in this section, to illustrate the range of other similar models models where learning and inference can be done tractably.

**Drawing a Joint Sample.** Given the auxiliary variable formulation, drawing a joint sample from a RC model is straightforward: pass messages inward to the root as before; compute the belief at the root, which is a distribution over global counts, and draw the value for the root variable from that distribution; now proceed outwards from the root towards the leaves. The one non-triviality is that values for the two children given a parent must be drawn simultaneously. To do this, first construct the belief at the trinary factor $f(z_p, z_l, z_r)$ conditioned on the value of $z_p$, which has been sampled already. Given $z_p$, there is a diagonal along the belief matrix corresponding to the values of $z_l$ and $z_r$ such that $z_l + z_r = z_p$. To draw the joint sample for $z_l$ and $z_r$, normalize this diagonal to sum to 1, then draw a value from that distribution. This gives the values for $z_l$ and $z_r$. Recurse downwards.

Other minor extensions appear in the Appendix.

### 5.4 Major Extension

Up until now, we have presented RC models as if the modeler who desires efficient exact inference must choose between standard pairwise trees or RC models. In fact, this is not necessary. It is possible to do exact inference in the following model in $\mathcal{O}(D \log^2 D)$ time:

$$p(\boldsymbol{y}) \propto \prod_{(d,d') \in \mathcal{E}} \theta_{dd'}(y_d, y_{d'}) \prod_{s_k \in \mathcal{S}(\mathcal{E})} f_k(\sum_{y_d \in s_k} y_d), \quad (8)$$

where $\mathcal{E}$ is an acyclic set of edges over variables $\boldsymbol{y}$, and $\mathcal{S}(\mathcal{E})$ is a set of nested subsets with subset structure that is "compatible" with $\mathcal{E}$. In other words, the modeler can choose the structure of either $\mathcal{E}$ or $\mathcal{S}(\mathcal{E})$ arbitrarily, and there is some non-degenerate choice of the other that allows for efficient inference, but not all combinations of trees and subset structures are compatible. Note that this structure cannot be represented by a standard Recursive Cardinality potential, because having e.g., edges $(d, d')$ and $(d', d'')$ would violate the nested subset structure requirement. The approach is similar to the base algorithm, but it involves constructing a junction tree that has separator sets involving one $y$ variable and one $z$ variable. We give details in the Appendix.

## 6 Experiments

In this section, we empirically explore the properties of the RC model, and demonstrate its usage in several interesting scenarios. Code will be made available implementing the convolution tree algorithm for computing marginals over $\boldsymbol{y}$ and $\boldsymbol{z}$, and for joint sampling.

### 6.1 Chain vs Tree as D grows

The version of the RC algorithm that we presented in Section 5 is the most efficient that we have discovered. There are, however, other approaches that are more efficient than $\mathcal{O}(D^3 \log D)$, but which are less efficient than the algorithm with the best asymptotic runtime. In this section, we compare the runtime of our algorithm (the "FFT Tree") against two baselines, both of which are improvements over the PL algorithm reviewed in Section 2.3.

The first baseline introduces auxiliary variables in a chain rather than tree structure. If we then follow the approach as in Section 5, this would lead to a $\mathcal{O}(D^2 \log D)$ algorithm. It turns out that in this case, messages can be computed via a summation of two arrays, so using FFTs is unnecessary. This yields a $\mathcal{O}(D^2)$ algorithm equivalent to that of Barlow and Heidtmann (1984), which we term the "Chain" algorithm. A drawback of this algorithm is that it also requires $\mathcal{O}(D^2)$ space. However, we note that if cardinalities greater than $k$ are disallowed by the cardinality potential, then this algorithm can be made to run in $\mathcal{O}(Dk)$ time, and it is perhaps the best choice.

The second baseline is the fast algorithm where FFTs are not used to compute messages, and instead we use an efficient but brute-force computation of the 1D convolutions required inside the message computations. We refer to this as the "Tree" algorithm. Solving the recurrence $T(n) = 2T(n/2) + n^2$ using the Master theorem, we see that the runtime is again $\mathcal{O}(D^2)$. The space usage is $\mathcal{O}(D \log D)$.

We run these algorithms on problems of size up to $2^{19}$ variables with a single random cardinality potential. The runtimes are reported in Fig. 3. The Chain algorithm fails after $D = 16000$ due to memory limits. The Tree algorithm is faster than the chain algorithm for larger $D$, but its quadratic time usage causes it to become quite slow once $D$ nears 100k. The FFT Tree algorithm behaves nearly linearly in practice, running on problems with half a million variables in less than 100 seconds. We note that in practice, for large values

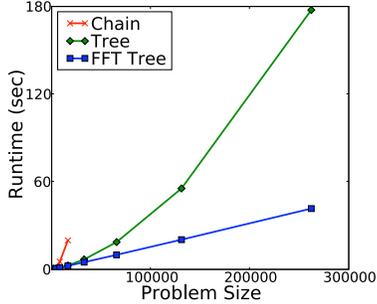

Figure 3: Runtimes of the different algorithms versus problem size. For very small $D$, the Chain algorithm is slightly faster, but it runs out of memory after around $D = 15000$, due to its quadratic memory usage. The FFT Tree algorithm runtime grows near linearly up through all experiments.

of $D$, care must be taken to avoid numerical issues for certain settings of model parameters.

### 6.2 Generalized Bipartite Matching

We have variables $\boldsymbol{y} = \{y_{ij}\}_{(i,j) \in D_I \times D_J}$, where $D_I$ is the number of rows, and $D_J$ is the number of columns. Here, we consider the Gibbs distribution defined by the following energy function:

$$E(\boldsymbol{y}) = \sum_{ij} \theta_{ij} y_{ij} + \sum_i f_c(\sum_j y_{ij}) + \sum_j f_r(\sum_i y_{ij}),$$

where $f_r$ and $f_c$ are functions of row and column counts. Note that if a constraint is placed on each row, saying that exactly one binary variable in each column can be on, then this formulation can also be used to represent cardinality potentials for multilabel problems. One motivation for this model comes from the problem of paper-to-reviewer matching, where $y_{ij}$ represents the event that paper $i$ is matched to reviewer $j$. The row and column functions can then be used to enforce the constraints that each paper should be given to e.g., 3 to 4 reviewers, and each reviewer should be assigned e.g., 6 to 8 papers. Note that the energy function can clearly represent the bipartite matching problem (by constraining row and column counts to be exactly 1), so computing marginals is #P-complete. Our approach will be to perform approximate inference using loopy belief propagation (LBP), where messages are computed using the FFT tree algorithm.

We compare against several baselines on small and medium-sized problems where we constrain columns to have 1 or 2 variables labeled 1, and rows are constrained to have 2 or 3 variables labeled 1. The first, "Node Marginals" is a naive approximation that simply ignores the constraints and computes marginals under the factorized distribution consisting of just the node potentials. The second is to find the exact MAP solution to the problem using a linear program (LP) and assume that the marginals are 0 or 1. The final baseline, which we take as the ground truth for the intractable medium-sized problem, is to use a block Gibbs sampler where blocks are chosen as the four variables $y_{i_1,j_1}, y_{i_1,j_2}, y_{i_2,j_1}, y_{i_2,j_2}$ for some choice of rows $i_1$ and $i_2$ and columns $j_1$ and $j_2$. To find a valid initial configuration, we initialize the sampler with the LP solution. We attempted multiple runs using random parameter settings. The results in Fig. 4 show that LBP achieves a low bias in a relatively short amount of time. For small problems, the constraints do not always greatly influence the marginals, as exhibited by the low bias of the naive approach in some runs; however, they clearly influence the larger problems. Finally, the LP method tends to be slower than LBP, becomes relatively slower on larger problems, and exhibits significant bias. We have run our algorithm on larger problems, and it converges quickly even on e.g., 100x100 problems, but we are unable to measure accuracy, because accurate baselines are prohibitively slow.

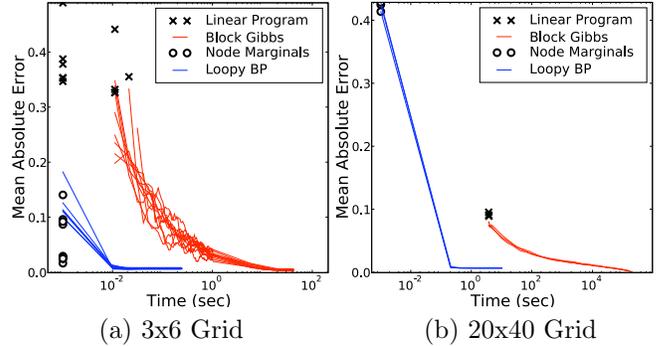

(a) 3x6 Grid   (b) 20x40 Grid

Figure 4: Mean absolute error between inferred marginals and ground truth marginals.

It is worth mentioning that this is a hard problem, and designing sampling schemes for many cases is nontrivial. Indeed, the Gibbs sampler we used will not be ergodic for disjoint cardinality constraints (e.g. only 3 or 10). By contrast, our method is relatively fast, accurate, and applies to a wide variety of constraints.

### 6.3 Multiple Instance Learning

In multiple instance learning (MIL), we are given "bags" of instances, which are labeled as either "positive" (at least one instance in the bag is positive) or "negative" (all instances in the bag are negative). This can be framed as a problem of learning with weak labels, where the weak labels take the form of a cardinality potential over individual instance labels. Indeed, MIL models where bag labels are modeled as noisy-OR can be seen as exactly this formulation, using the form of noisy-OR from Eq. (5). However, given the fast algorithms for Recursive Cardinality potentials developed in this work, it becomes tractable to assert other forms of the distribution over within-bag counts. In this section, we experiment with this alternative.

More formally, for variables $\boldsymbol{y}$ appearing in a bag labeled as $t = 0$ (negative) or $t = 1$ (positive), we can rewrite the likelihood of a label as $\sum_{\boldsymbol{y}} p(t, \boldsymbol{y})$, where

$$p(t, \boldsymbol{y}) \propto f^{(t)}(\sum_d y_d) \prod_d \exp\{\theta_d \cdot y_d\}, \qquad (9)$$

where $f^{(t)}$ is some cardinality function that imposes a preference on the number of binary variables turned on for bags labeled as $t$, and $\theta_d$ is a unary potential. All of the required quantities for learning can be computed in two calls to our algorithm (corresponding to $t = 0$ and $t = 1$).

The standard data set for evaluating MIL learners is the Musk dataset from Dietterich et al. (1997). In the data, bags molecules are labeled as to whether they are "musks." A molecule is considered to be a musk if any low energy conformations of the molecule is a musk. The bags in this data set correspond to molecules, and the instances are features of many different low energy conformations. We compared two methods on the musk1 version of the dataset: a standard Noisy-OR model, and a "Normal" model, where $f^{(0)}(c) = \exp\{-(\frac{c}{D})/2\sigma^2\}$ and $f^{(1)}(c) = \exp\{-(\mu - \frac{c}{D})/2\sigma^2\}$.

As is common in the MIL literature, we divide the data into 10 evenly-sized folds and run 10-fold cross-validation. We use 20% of bags for validation, and 10% for testing in each split. To set an $L1$ regularization parameter, and to choose the $\sigma$ and $\epsilon$ parameters in the Normal and Noisy-OR models, respectively, we do a grid search and choose the setting that produced the best average validation error across folds. Fig. 5 (a) reports results showing how error varies as a function of the $\mu$ and $\lambda$ parameters for Normal and Noisy-OR, respectively. We note that the musk1 dataset has only 92 instances, and the standard deviation of errors across validation folds was high, so these results are not statistically significant. However, we do see a trend where varying the $f$ functions can affect performance, and that a Normal cardinality potential out-performs the standard Noisy-OR. For this problem, it appears that setting the $\mu$ parameter to be large is beneficial for learning, but in general this should be a parameter that is set via prior knowledge or cross validation. Also note that Gehler and Chapelle (2007) enforced a similar bias on bag counts within an SVM formulation, and their results also suggested that encouraging the model to turn on more binary variables within positive bags improved performance on the musk1 data. In Fig. 5 (b), we show how parameter values in the cardinality potential affects the expected number of instances labeled as "on" within positive bags. The trend in this plot shows that varying $f$ does indeed affect how many instances within a positive bag that the model chooses to label as positive examples.

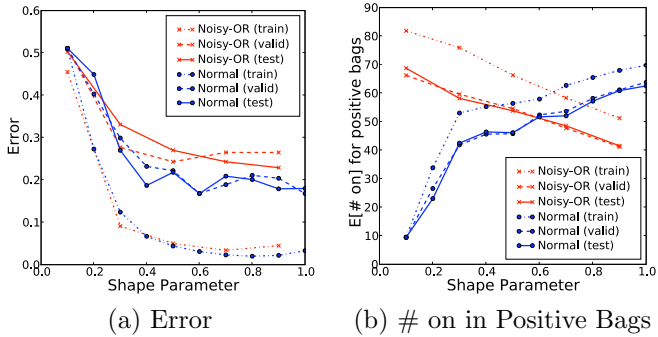

(a) Error  (b) # on in Positive Bags

Figure 5: Multiple instance learning results.

### 6.4 Generative Image Models

In this set of experiments, we attempt to distinguish between the representational abilities of pairwise tree-structured graphical models, and the RC model. We consider a synthetic dataset that has been sampled from an Ising model near the critical temperature resulting in images with highly correlated regions. The pairwise edges form a grid structure, and the resulting data consists of 1000 $16 \times 16$ images. In order to compare the pairwise tree and the RC models we first learn their parameters on each dataset using nonlinear conjugate gradients to minimize the negative log-likelihood. After learning each model, we generate the same number of samples as the original dataset and use these to compare statistics. After learning, the log-likelihoods for the pairwise tree model was 145.22, while the log-likelihoods for the RC model was 146.78. We hypothesize that this difference is because the pairwise tree model is able to do a better job at directly capturing pairwise statistics, however the RC model is better at capturing long-range, higher-order statistics. We demonstrate this in our experiments below.

**Heuristic choice of transformation structure.** For each dataset, in order to determine the structure of the nested cardinality potential, we run hierarchical agglomerative clustering on images, where the similarity between pixels is determined by how often they take the same label. This gives us a full tree structure over the pixels, which we then directly use to give the nested subset structure. We leave a study of other ways for choosing trees to future work.

**Pairwise statistics.** We analyze the pairwise statistics of the learned distributions in comparison to the original datasets. Fig. 6 shows the pairwise correla-tions between each pixel, with lighter shading corresponding to stronger correlation. The Ising model enforces strong correlation between neighbouring pixels, and this is reflected in the banded nature of the correlation matrix. Visually, we can see that the RC model does a better job of capturing long range correlations corresponding to the more faded bands away from the main diagonal.

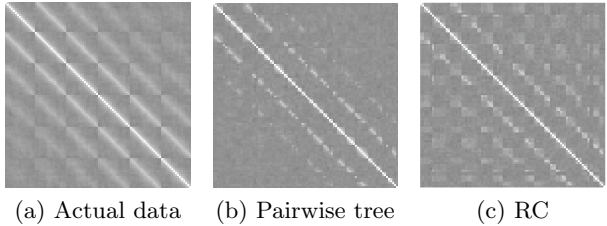

(a) Actual data  (b) Pairwise tree  (c) RC

Figure 6: Zoomed pairwise correlation matrices from the Ising dataset.

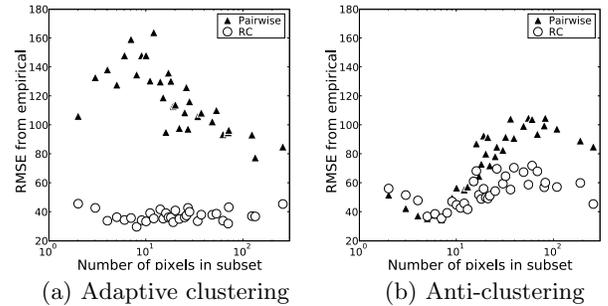

(a) Adaptive clustering  (b) Anti-clustering

Figure 7: Error in higher-order count statistics between the empirical and model distributions as a function of the number of variables in a subset.

**Higher-order statistics.** As a final comparison, we analyze the ability of both models to capture higher-order statistics in the data. Specifically, we test the ability of the RC model to capture higher-order count statistics over subsets that it does not directly encode. To do this, we first build a nested tree over the variables using hierarchical clustering. Using this tree, we look at the count distributions of subsets and compare the empirical and model distributions. In Fig. 7, we show the root mean squared error between the empirical and model distributions as a function of the number of variables in a subset for the Ising dataset. We compare both models on nested trees constructed from two heuristics: the first is the same scheme we use to parameterize the RC model, which we call adaptive clustering. For the second, we attempt to create a "worst case" clustering by taking the distance values used for the adaptive method and negating them before clustering; we call this scheme anti-clustering. We also tried random clusterings, however the results were not qualitatively different from the anti-clustering scheme. Our results show that the RC model does a better job than the pairwise tree model at capturing higher-order count statistics, even if it does not encode them directly. Particularly interesting is that for the adaptive tree, the pairwise model does significantly worse than the RC model. This is due to the confluence of two effects: first, the hierarchical clustering that determines the RC structure seeks to choose subsets of variables that are highly correlated to merge into a subset; second, the spanning tree is unable to model this "all-or-none" effect, and instead tends towards a binomial-like distribution with mode near the 50% full point. This leads to large errors.

## 7  Discussion

We have presented a new class of high order model where exact inference can be done very efficiently. This model can be used on its own, and it is already able to capture a mix of high and low order statistics that would be difficult for other tractable models. The model can also be used as a subcomponent within outer approximate inference procedures, allowing efficient approximate marginal inference for hard problems that few other approaches are able to tackle.

The general approach—of adding a set of auxiliary variables with highly structured interactions so as to guarantee efficient message passing inference—leads to efficient exact inference in other extended models such as tree structured models augmented with a cardinality potential, and we believe the approach to extend even beyond that.

Finally, we have presented a diverse set of applications and shown how many problems that are not naturally thought of in terms of cardinality potentials can be cast as such. We believe that the highly efficient algorithms discussed here will only increase the number of applications where cardinality potentials are useful.

## References

Barlow, R. E., & Heidtmann, K. D. (1984). Computing k-out-of-n system reliability. *IEEE Transactions on Reliability*, *33*, 322-323.

Belfore, L. (1995). An O(n) log2(n) algorithm for computing the reliability of k-out-of-n:G and k-to-l-out-of-n:G systems. *IEEE Transactions on Reliability*, *44*(1).

Chen, S. X., & Liu, J. S. (1997). Statistical applications of the Poisson-Binomial and conditional Bernoulli distributions. *Statistica Sinica*, *7*(4).

de Salvo Braz, R., Amir, E., & Roth, D. (2005). Lifted

first-order probabilistic inference. In *In proceedings of ijcai-05, 19th international joint conference on artificial intelligence* (pp. 1319–1325). Morgan Kaufmann.

Dietterich, T. G., Lathrop, R. H., Lozano-Perez, T., & Pharmaceutical, A. (1997). Solving the multiple-instance problem with axis-parallel rectangles. *Artificial Intelligence*, *89*, 31–71.

Felzenszwalb, P. F., & Huttenlocher, D. P. (2004). Efficient belief propagation for early vision. In *CVPR* (pp. 261–268).

Fernandez, M., & Williams, S. (2010). Closed-form expression for the poisson-binomial probability density function. *IEEE Transactions on Aerospace Electronic Systems*, *46*, 803-81.

Fisher, M. E. (1961, Dec). Statistical mechanics of dimers on a plane lattice. *Phys. Rev.*, *124*, 1664–1672. Available from http://link.aps.org/doi/10.1103/PhysRev.124.1664

Gail, M. H., Lubin, J. H., & Rubinstein, L. V. (1981). Likelihood calculations for matched case-control studies and survival studies with tied death times. *Biometrika*.

Gallager, R. (1963). *Low-density parity-check codes*. MIT Press.

Ganchev, K., Graça, J., Gillenwater, J., & Taskar, B. (2010). Posterior regularization for structured latent variable models. *Journal of Machine Learning Research*, *11*, 2001-2049.

Gehler, P., & Chapelle, O. (2007, March). Deterministic annealing for multiple-instance learning. In Meila, M., & X. Shen (Eds.), *ICML* (p. 123-130).

Gould, S. (2011). Max-margin learning for lower linear envelope potentials in binary markov random fields. In *ICML*.

Gupta, R., Diwan, A., & Sarawagi, S. (2007). Efficient inference with cardinality-based clique potentials. In *ICML* (Vol. 227, p. 329-336).

Hong, Y. (2011). *On computing the distribution function for the sum of independent and non-identical random indicators.*

Huang, B., & Jebara, T. (2007). Loopy belief propagation for bipartite maximum weight b-matching. In *AISTATS*.

Kohli, P., Ladicky, L., & Torr, P. H. S. (2009). Robust higher order potentials for enforcing label consistency. *International Journal of Computer Vision*, *82*(3), 302-324.

Marlin, B., Swersky, K., Chen, B., & Freitas, N. de. (2010). Inductive principles for restricted boltzmann machine learning. In *AISTATS*.

Mezuman, E., & Weiss, Y. (2012). Globally optimizing graph partitioning problems using message passing. In *AISTATS*.

Milch, B., Zettlemoyer, L. S., Kersting, K., Haimes, M., & Kaelbling, L. P. (2008). Lifted probabilistic inference with counting formulas. In D. Fox & C. P. Gomes (Eds.), *Aaai* (p. 1062-1068). AAAI Press.

Potetz, B., & Lee, T. S. (2008, Oct). Efficient belief propagation for higher order cliques using linear constraint nodes. *Computer Vision and Image Understanding*, *112*(1), 39–54.

Ramalingham, S., Kohli, P., Alahari, K., & Torr, P. (2008). Exact inference in multi-label CRFs with higher order cliques. In *CVPR*.

Régin, J.-C. (1996). Generalized arc consistency for global cardinality constraint. In *AAAI/IAAI* (p. 209-215).

Rother, C., Kohli, P., Feng, W., & Jia, J. (2009). Minimizing sparse higher order energy functions of discrete variables. In *CVPR* (p. 1382-1389).

Stobbe, P., & Krause, A. (2010). Efficient minimization of decomposable submodular functions. In *NIPS*.

Tarlow, D., Givoni, I., & Zemel, R. (2010). HOP-MAP: Efficient message passing for high order potentials. In *AISTATS*.

Vicente, S., Kolmogorov, V., & Rother, C. (2009). Joint optimization of segmentation and appearance models. In *ICCV*.

Zhao, P., Rocha, G., & Yu, B. (2006). Grouped and hierarchical model selection through composite absolute penalties. *Department of Statistics, UC Berkeley, Tech. Rep*, *703*.